\PassOptionsToPackage{table}{xcolor}
\documentclass[runningheads]{llncs}
\usepackage[T1]{fontenc}

\usepackage{iciap}

\usepackage{iciapabbrv}
\usepackage[accsupp]{axessibility}

\usepackage{hyperref}

\usepackage{orcidlink}   
\usepackage{graphicx}
\usepackage{caption}
\usepackage{booktabs}
\usepackage{amsmath}
\usepackage{amssymb}
\usepackage{soul}
\usepackage{color}

\begin{document}

\title{Efficient Calisthenics Skills Classification through 
Foreground Instance Selection and Depth Estimation}

\titlerunning{Efficient Calisthenics Skills Classification}

\author{Antonio Finocchiaro\textsuperscript{1} \and  
Giovanni Maria Farinella\textsuperscript{1}\orcidID{0000-0002-6034-0432} \and  
Antonino Furnari\textsuperscript{1}\orcidID{0000-0001-6911-0302}  
}

\authorrunning{Finocchiaro et al.}  

\institute{\textsuperscript{1} Department of Mathematics and Computer Science, University of Catania, Italy \\  
\email{1000006272@studium.unict.it, \{giovanni.farinella, antonino.furnari\}@unict.it} \\  
\url{https://iplab.dmi.unict.it/fpv/}  
}

\maketitle
\begin{abstract}
Calisthenics skill classification is the computer vision task of inferring the skill performed by an athlete from images, enabling automatic performance assessment and personalized analytics. Traditional methods for calisthenics skill recognition are based on pose estimation methods to determine the position of skeletal data from images, which is later fed to a classification algorithm to infer the performed skill. Despite the progress in human pose estimation algorithms, they still involve high computational costs, long inference times, and complex setups, which limit the applicability of such approaches in real-time applications or mobile devices. This work proposes a direct approach to calisthenics skill recognition, which leverages depth estimation and athlete patch retrieval to avoid the computationally expensive human pose estimation module. Using Depth Anything V2 for depth estimation and YOLOv10 for athlete localization, we segment the subject from the background rather than relying on traditional pose estimation techniques. This strategy increases efficiency, reduces inference time, and improves classification accuracy. Our approach significantly outperforms skeleton-based methods, achieving 38.3$\times$ faster inference with RGB image patches and improved classification accuracy with depth patches (0.837 vs. 0.815). Beyond these performance gains, the modular design of our pipeline allows for flexible replacement of components, enabling future enhancements and adaptation to real-world applications.
\keywords{Calisthenics Skill Recognition \and Pose Classification  \and Foreground Instance Selection \and Depth Estimation.}
\end{abstract}
\section{Introduction}
Calisthenics \cite{low2016} is a bodyweight discipline with categories including skills, endurance, and streetlifting. The skills category features complex movements requiring strength, stability, and coordination. In competitions, athletes perform sequences combining static holds and dynamic transitions, with evaluation based on skill execution accuracy and duration.
In this work, we focus on static skills, which are poses held by the athlete for several seconds, requiring significant muscular control and balance. Accurate recognition of these skills is critical, especially in scenarios involving multiple people, such as crowded environments or varying camera perspectives. To accomplish this task, typical approaches leverage pose estimation to extract key points from athletes' poses, focusing on body movement and alignment while reducing background noise. However, this methodology is computationally expensive, requiring approximately 0.383 seconds per inference, making it impractical for real-world applications. Additionally, it struggles with joint occlusions and is sensitive to lighting and complex backgrounds.

\begin{figure}[t]
\includegraphics[width=\textwidth]{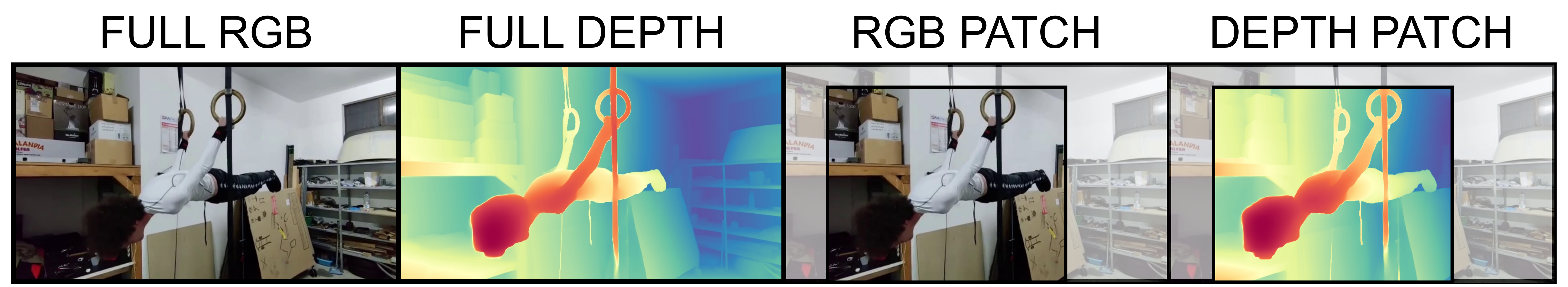}
\caption{Calisthenics skill classification can be done considering different input modalities. While a depth image helps discriminating the athlete from the background with respect to the RGB frame, we found the best results when extracting patches of the athlete from the RGB or depth image.} 
\label{fig:intro}
\end{figure}

To overcome these limitations, we investigate four computationally efficient approaches which do not require human pose estimation, the main computational bottleneck. The first approach classifies the skill directly from a full RGB frame using a Convolutional Neural Network (CNN). While this approach is computationally efficient, it is affected by the background noise common in sports footage. Hypothesizing that depth could help isolate athletes and reduce clutter, we propose a second methodology which processes depth images extracted using Depth Anything V2 (DAV2) \cite{yang2024depthv2} with a CNN. To further isolate the background from the foreground instance, we integrated YOLOv10 \cite{wang2024yolov10realtimeendtoendobject} to detect the athlete in the images. Finally, we combined both methods and estimated the depth from the extracted RGB patches and then fed them to the CNN for classification.
Figure \ref{fig:intro} reports examples of the input modalities considered by the proposed approaches, which are subsequently
tested and compared against a pose-based skill classification approach \cite{finocchiaro2024calisthenics}, demonstrating superior efficiency and accuracy.

\section{Related Works}
In recent years, computer vision-based techniques have played a crucial role in sports analysis, enabling the development of automated tools for match analysis, performance evaluation, and referee assistance. 

\noindent \textbf{Computer Vision in Sports Analysis}
Different previous works considered the application of Computer Vision for automated sports analysis. Pioneering approaches for Olympic event scoring \cite{parmar2017learning} and frameworks for action quality assessment across multiple athletic activities \cite{parmar2019action} established metrics for evaluating complex performances. In soccer analysis, methods for classifying ball-on-goal position through kicker shooting action \cite{toron2023classifying} demonstrated how vision-based systems provide tactical insights. Research expanded to endurance sports with contributions on ultra-distance running using I3D ConvNet transfer learning \cite{freire2022towards} and large-scale analysis of athletes' cumulative race times \cite{freire2023large}. In addition, the challenges of re-identifying athletes in sports scenarios were addressed in \cite{freire2023largev2}.
The research on action recognition in calisthenics has been recently investigated in \cite{finocchiaro2024calisthenics}, which proposes a new dataset and an approach based on human pose estimation. We investigate computationally efficient approaches to calisthenics skill classification based on the direct processing of RGB or depth frames.

\noindent
\textbf{Skeleton-Based Approaches for Action Recognition}
Skeleton-based pose estimation relies on detecting key points representing human joints to analyze motion and infer activity. Several models \cite{cao2019openposerealtimemultiperson2d, güler2018denseposedensehumanpose, jiang2023rtmposerealtimemultipersonpose} have demonstrated high accuracy in joint detection for both single and multi-person scenarios. Recent advances incorporate temporal information through Spatial-Temporal Graph Convolutional Networks (ST-GCN) \cite{yan2018spatialtemporalgraphconvolutional} and attention-enhanced variants \cite{si2019attentionenhancedgraphconvolutional}, achieving superior performance in sports with complex movements. While \cite{finocchiaro2024calisthenics} addressed calisthenics skill classification using a skeleton-based approach, we propose an RGB-based method using full-image representations instead of explicit body joint localization.

\noindent
\textbf{RGB-Based Approaches for Action Recognition}
In contrast to skeleton-based methods, RGB-based approaches directly use full-frame information for activity detection and pose estimation processes.
Recently, depth-based pose classification systems have shown significant improvements over traditional RGB-based methods \cite{efficientposeestsingledepth, 3dhpestdepthgolf, depth-img-forest, accura-only, app9122478}. Advancements in monocular depth estimation \cite{ranftl2020towards} have enabled robust cross-dataset transfer learning without specialized depth sensors. These approaches improve robustness against background variations that introduce noise into classification models. Similarly, we perform calisthenics skill classification directly from RGB and depth inputs, showing that foreground instance selection reduces background clutter influence.

\noindent
\textbf{Influence of Background in Image Recognition}
In image-based recognition tasks, the relationship between foreground objects and background elements plays a critical role in classification performance. Several studies have explored the correlation between object recognition and background context, including \cite{bhatt2023mitigatingeffectincidentalcorrelations, chou2023finegrainedvisualclassificationhightemperature, aniraj2023maskingstrategiesbackgroundbias, wang2022cladcontrastivelearningbased}.
While background-aware learning mechanisms often benefit general image recognition, research in Human Activity Recognition (HAR) \cite{freire2022towards, freire2023large} presents approaches that neutralize backgrounds through averaging rather than elimination, suggesting background context may contain valuable information for certain tasks. In contrast, we investigate complete background elimination through foreground instance selection. This work focuses on skill classification from images, where background influence is removed by detecting the athlete and using depth estimation. We evaluate classification performance in two settings: using the full image versus only foreground information.



\section{Investigated Approaches}
In this work, we aim to enhance the performance of calisthenics skill classification using full RGB images as input, without any prior information about the localization of human body joints. To achieve this, we evaluate four different approaches that work with different types of input data (see Figure \ref{fig:intro}), using DAV2 for depth estimation and YOLOv10 for the patch extraction process.
\begin{figure}[t]
\includegraphics[width=\textwidth]{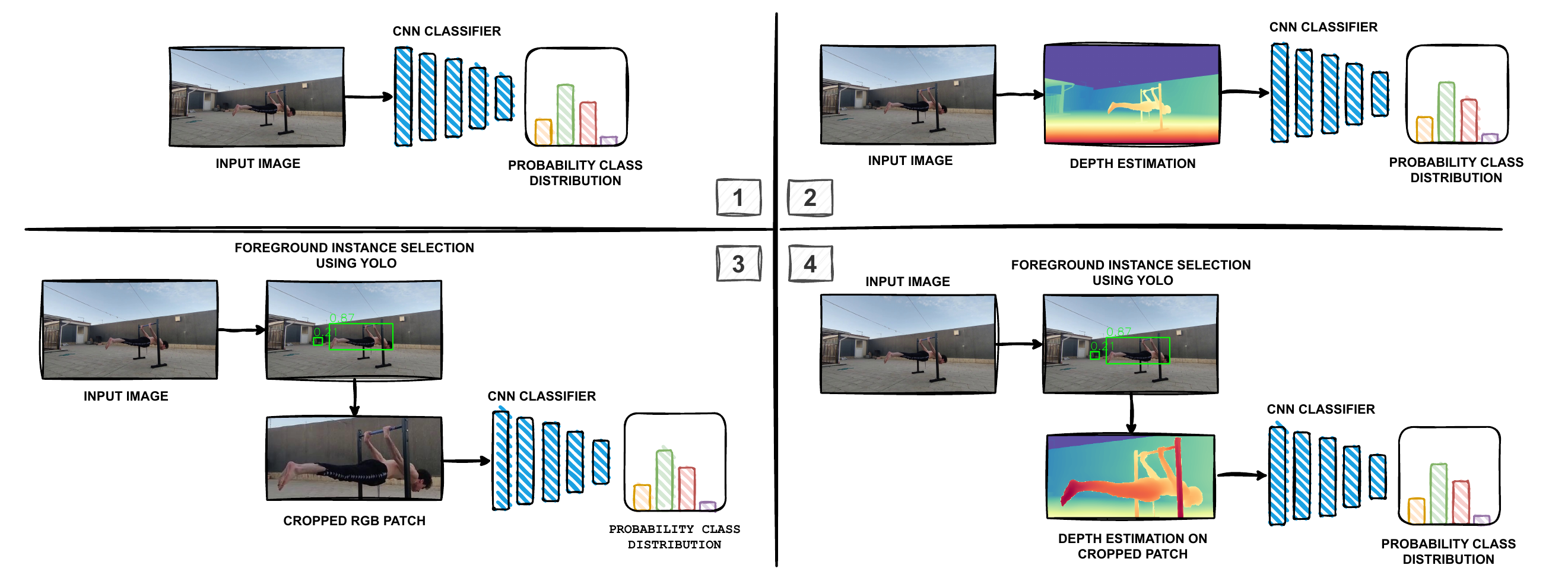}
\caption{Diagrams of the approaches investigated in this work. [1] Approach 1: Full RGB image classification, [2] Approach 2: Full depth image classification, [3] Approach 3: RGB patch classification, [4] Approach 4: Depth patch classification.} 
\label{fig:approaches}
\end{figure}

Figure \ref{fig:approaches} illustrates the various approaches, which we summarized in the following and detailed in the next sections.

\begin{itemize}
    \item \textbf{Approach 1} consists of feeding the full RGB image directly into a CNN for classification, which serves as a baseline approach without any background noise removal mechanism.
    \item \textbf{Approach 2} introduces a depth estimation model \cite{yang2024depthv2} to filter out background noise, exploiting the assumption that the athlete is closer to the camera, and then feeds the depth image to a CNN for classification.
    \item \textbf{Approach 3} performs foreground instance selection through the Object Detection (OD) model \cite{wang2024yolov10realtimeendtoendobject} to extract an RGB image patch representing the athlete, which the CNN then classifies.
    \item \textbf{Approach 4} leverages the OD model \cite{wang2024yolov10realtimeendtoendobject} to first extract the athlete’s RGB patch, and then applies depth estimation with \cite{yang2024depthv2} to this localized region. The resulting depth patch is then used as input to the CNN for classification.

\end{itemize} 

\subsection{Details on the adopted Components}
\begin{figure}[t]
\includegraphics[width=\textwidth]{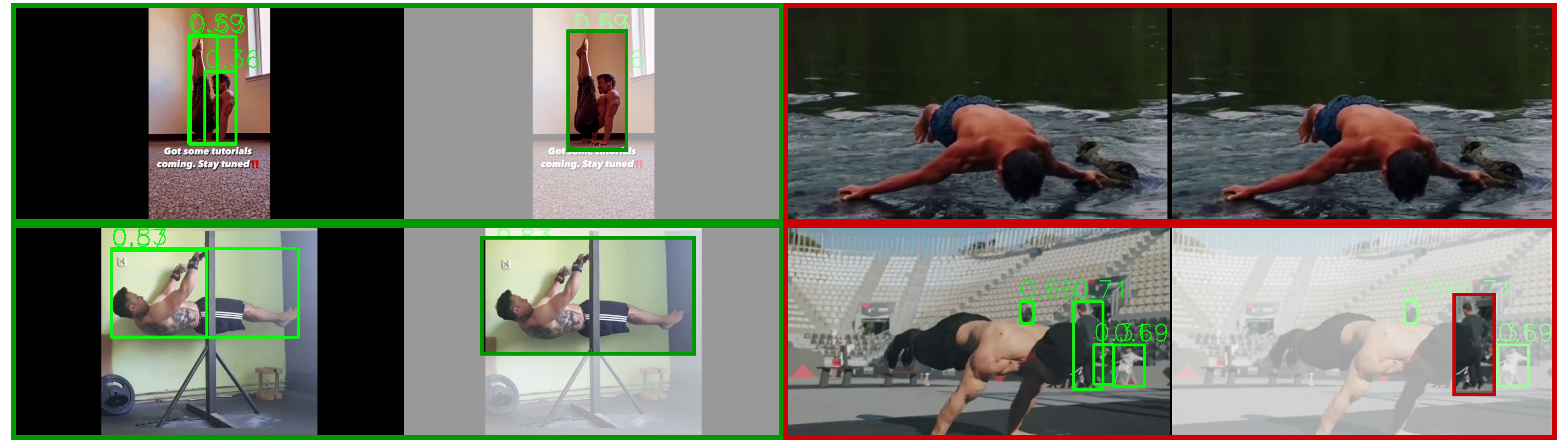}
\caption{Representative samples of patch extraction post-processing. Green boxes indicate successful retrieval cases, while red boxes highlight failures—top: no athlete detected, bottom: incorrect person detected. All extracted patches demonstrate the bounding box area enlargement mechanism.} 
\label{fig:yolo-post}
\end{figure}

This section discusses the details of the adopted components.
\begin{itemize}
    \item \textbf{Depth Estimation} is obtained using the original version of DAV2 presented in \cite{yang2024depthv2}, pretrained on 595K synthetic labeled images and 62M+ real unlabeled images. Specifically, we used the \textit{ViTs} encoder for inference and saved the resulting depth images with an applied colormap.
    \item \textbf{Feature Extraction and Multi-class Skill Classification} task is tackled by training and evaluating a simple CNN. Specifically, we tested: ResNet101 \cite{he2015deepresiduallearningimage}, ResNeXt101 \cite{xie2017aggregatedresidualtransformationsdeep}, MobileNetV3 \cite{howard2017mobilenetsefficientconvolutionalneural}, EfficientNetV2 \cite{tan2021efficientnetv2smallermodelsfaster}, and InceptionV3 \cite{szegedy2015rethinkinginceptionarchitecturecomputer} which offer different trade-off between performance and computational overhead. For each network, we considered both the randomly initialized weight version and the pretrained one, trained on ImageNet1K, a subset of ImageNet \cite{russakovsky2015imagenetlargescalevisual}. 
    \item \textbf{Foreground Instance Selection using YOLO} was handled by a pretrained YOLOv10 model\footnote{\url{https://www.kaggle.com/code/cubeai/person-detection-with-yolov10/output}}, which performs person detection and is used as inference to identify the athlete in each frame. This approach enables extracting athlete patches rather than using the full image.
        
    To select the most prominent athlete in the case of multiple detections, each bounding box is assigned a score obtained by a weighted average between the detection confidence score (weight 0.6) and the bounding box area normalized by the image size (weight 0.4). Finally, we select the box with the highest score.
    When no bounding box was detected or when the detected athlete patch occupies less than 1\% of the frame area, we apply a center crop that maintains the original aspect ratio but is 20\% smaller than the full frame. This helps avoid detecting the wrong people in the scene who might be in the background, thereby removing false positives. For valid detections, we apply a dynamic enlargement mechanism where smaller boxes receive greater proportional enlargement (up to 15\%) and larger boxes receive minimal enlargement (at least 5\%), with the exact factor inversely related to the box's area ratio. The coordinates are then clipped to fit within image boundaries. With this refinement, we aim to add robustness to the detection module, ensuring that athletes remain fully visible in the extracted patches regardless of their relative size in the frame. The resulting patches were resized to \(224 \times 224\) and fed into the CNN. Figure \ref{fig:yolo-post} illustrates the process of selecting the most prominent athlete's bounding box. It highlights both successful and erroneous cases, where misidentifying the athlete propagates through the pipeline, leading to incorrect pose classification.

\end{itemize}

\section{Experimental Settings and Results}

In this section, we refer to the approaches following the same order indicated in Figure \ref{fig:approaches}. The performance of the CNNs will be analyzed using approaches 1 \& 2 on the full images. Then, the CNN with the best performance will be used to evaluate approaches 3 \& 4. The dataset used to classify the skills is the one proposed in \cite{finocchiaro2024calisthenics}, provided by the authors for research purposes. To our knowledge, it is currently the only available dataset of calisthenics skills.
The dataset comprises 839 frame-wise annotated videos of athletes performing calisthenics skills, processed at 960×540 and 24 fps. We extracted every frame, resulting in 115,843 frames capturing skill executions, transitions, and random movements retained to help discriminate the background class.
The dataset includes nine skills — BL, FL, FLAG, IC, MAL, OAFL, OAHS, PL, and VSIT — and an additional background class (NONE).
To account for perspective changes, we applied transformations such as horizontal flipping and motion blur.
The dataset used to train the OD\footnote{https://www.kaggle.com/datasets/cubeai/person-detection-for-yolov8/data} contains 13,754 training and 4,000 validation images featuring people engaged in different activities. While not specifically for calisthenics, the domain gap is not a concern as the task remains person detection.
The object detector, YOLOv10n (Nano), was trained for 100 epochs using SGD with learning rate 0.01, momentum 0.9, and batch size 16. The input size was 640x640px with padding to preserve aspect ratio. A 0.2 confidence threshold was set to minimize missed detections.
For classification, each CNN was trained for 100 epochs with custom early stopping (0.001 tolerance, 15 epochs patience) using Focal Loss, Adam optimizer with OneCycleLR scheduler (maximum 0.01), Softplus activation function, and batch size 128. The optimal hyperparameters were determined by Bayesian optimization \cite{frazier2018tutorialbayesianoptimization}.
All experiments were performed on a \textit{Nvidia Tesla V100S 32GB GPU}.


\subsection{Comparison Between Various CNNs}
\begin{table}[t]
\centering
\resizebox{\columnwidth}{!}{%
\begin{tabular}{lcccccc}
\toprule
\textbf{Model} & \textbf{Pretrained} & \textbf{Input} & \textbf{Precision $\uparrow$} & \textbf{Recall $\uparrow$} & \textbf{F1 Score $\uparrow$} & \textbf{Accuracy $\uparrow$} \\ 
\midrule
ResNet101      & No  & Full RGB   & 0.532 & 0.409 & 0.437 & 0.429 \\
ResNeXt101     & No  & Full RGB   & 0.463 & 0.336 & 0.367 & 0.371 \\
MobileNetV3    & No  & Full RGB   & 0.407 & 0.366 & 0.370 & 0.362 \\
EfficientNetV2 & No  & Full RGB   & \textbf{0.552} & \textbf{0.510} & \textbf{0.523} & \textbf{0.519} \\
InceptionV3    & No  & Full RGB   & 0.410 & 0.351 & 0.358 & 0.357 \\
\midrule

ResNet101      & Yes   & Full RGB   & 0.731 & 0.686 & 0.702 & 0.699 \\
ResNeXt101     & Yes   & Full RGB   & 0.743 & 0.667 & 0.696 & 0.700 \\
MobileNetV3    & Yes   & Full RGB   & 0.707 & 0.651 & 0.671 & 0.666 \\
EfficientNetV2 & Yes   & Full RGB   & \textbf{0.753} & \textbf{0.734} & \textbf{0.734} & \textbf{0.726} \\
InceptionV3    & Yes   & Full RGB   & 0.750 & 0.676 & 0.704 & 0.698 \\

\midrule

ResNet101      & No  & Full Depth & 0.758 & 0.708 & 0.723 & 0.703 \\
ResNeXt101     & No  & Full Depth & \textbf{0.767} & 0.699 & 0.723 & 0.696 \\
MobileNetV3    & No  & Full Depth & 0.700 & 0.673 & 0.677 & 0.671 \\
EfficientNetV2 & No  & Full Depth & 0.740 & \textbf{0.716} & \textbf{0.724} & \textbf{0.710} \\
InceptionV3    & No  & Full Depth & 0.694 & 0.618 & 0.643 & 0.630 \\
\midrule

ResNet101      & Yes   & Full Depth & \textbf{0.808} & 0.762 & 0.780 & 0.760 \\
ResNeXt101     & Yes   & Full Depth & 0.796 & 0.773 & 0.780 & 0.766 \\
MobileNetV3    & Yes   & Full Depth & 0.789 & 0.731 & 0.754 & 0.744 \\
EfficientNetV2 & Yes   & Full Depth & 0.793 & \textbf{0.789} & \textbf{0.787} & \textbf{0.778} \\
InceptionV3    & Yes   & Full Depth & 0.793 & 0.771 & 0.780 & 0.769 \\
\bottomrule
\end{tabular}%
}

\caption{Comparison of CNN classification models. Pretrained CNNs have been previously trained on ImageNet1K \cite{russakovsky2015imagenetlargescalevisual}.
For each row, we evaluated the model using Precision, Recall, F1 Score, and Accuracy. \textbf{Bold} values indicate the best-performing metric for each configuration (model, pretrained/non-pretrained and input type) across all evaluation metrics.}
\label{tab:quantres}
\end{table}

Table \ref{tab:quantres} summarizes the performance of the CNN classification models in terms of precision, recall, F1 score, and accuracy. Each experiment was conducted using both randomly initialized weights, where the entire CNN was trained from scratch, and fine-tuned weights pretrained on ImageNet1K\cite{russakovsky2015imagenetlargescalevisual}. This approach was taken to evaluate the impact of transfer learning, using pretrained features to improve convergence speed and generalization. 

EfficientNetV2 shows the highest performance in most configurations, with the best results achieved by pretrained models using depth images. When comparing full depth images to RGB images in non-pretrained settings, the accuracy of ResNet101 improves from 0.429 to 0.703 ($+0.274$), ResNeXt from 0.371 to 0.696 ($+0.325$), MobileNetV3 from 0.362 to 0.671 ($+0.309$), EfficientNetV2 from 0.519 to 0.710 ($+0.191$), and InceptionV3 from 0.357 to 0.630 ($+0.273$). The highest accuracy (0.778) is achieved by EfficientNetV2 using depth images and pretrained weights. Based on its performance, we adopt EfficientNetV2 for subsequent experiments. These results show consistent accuracy gains for all models with depth maps, indicating that they provide more discriminative features than RGB images.

\subsection{Patch-based skill classification}

\begin{table}[t]
\centering
\resizebox{\textwidth}{!}{%
\begin{tabular}{lcccccc}
\toprule
\textbf{Model}          & \textbf{Pretrained} & \textbf{Input} & \textbf{Precision $\uparrow$} & \textbf{Recall $\uparrow$} & \textbf{F1 Score $\uparrow$} & \textbf{Accuracy $\uparrow$} \\ 
\midrule
EfficientNetV2         & No  & Full RGB         & 0.552 & 0.510 & 0.523 & 0.519 \\
EfficientNetV2         & No  & RGB Patches  & \textbf{0.656} & \textbf{0.611} & \textbf{0.629} & \textbf{0.626} \\
\midrule

EfficientNetV2         & Yes   & Full RGB         & 0.753 & 0.734 & 0.734 & 0.726 \\
EfficientNetV2         & Yes   & RGB Patches  & \textbf{0.801} & \textbf{0.787} & \textbf{0.790} & \textbf{0.792} \\
\midrule

EfficientNetV2         & No  & Full Depth       & 0.740 & 0.716 & 0.724 & 0.710 \\
EfficientNetV2         & No  & Depth Patches & \textbf{0.849} & \textbf{0.837} & \textbf{0.838} & \textbf{0.837} \\
\midrule

EfficientNetV2         & Yes   & Full Depth       & 0.793 & 0.789 & 0.787 & 0.778 \\
EfficientNetV2         & Yes   & Depth Patches & \textbf{0.838} & \textbf{0.830} & \textbf{0.831} & \textbf{0.834} \\
\bottomrule
\end{tabular}%
}
\caption{Comparison between all EfficientNetV2 configurations. Pretrained CNNs have been previously trained on ImageNet1K \cite{russakovsky2015imagenetlargescalevisual}.
Each row presents Precision, Recall, F1 Score, and Accuracy. \textbf{Bold} values indicate the best-performing metric for each configuration, comparing the full input experiment with its corresponding patches across all evaluation criteria.}
\label{tab:effnet_performance}
\end{table}

Table \ref{tab:effnet_performance} presents experiments using both full and patch-based inputs for RGB and depth images, with non-pretrained and pretrained models. 
Of the 115,843 frames in the dataset, only 3,017 frames (2.60\%) required center cropping due to absent detections or detections with areas smaller than 1\% of the frame size, indicating the detection algorithm successfully identified meaningful regions of interest in most frames.
For RGB inputs, patches consistently outperformed full images. In the non-pretrained case, accuracy increased from 0.519 to 0.626, while in the pretrained setting, it improved from 0.726 to 0.792.
For depth inputs, patches also showed superior performance. Without pretraining, accuracy improved from 0.710 to 0.837, and with pretraining, from 0.778 to 0.834. These results demonstrate the effectiveness of integrating depth information into localized patches for classification.


\subsection{Analysis on Inference Time}
In this section, we evaluate the inference time of the different approaches to determine the most appropriate method based on performance requirements and to identify potential bottlenecks.

We consider ``1 CS IIT'' to be the time in seconds required for a single cold start inference, which is significantly higher due to model initialization, while the column ``10 CS ITT" refers to the time in seconds needed to perform inference on 10 images. 
``AVG IIT" 
represents the average time in seconds it takes to make an inference after making the first one. 

To determine the best trade-off between model accuracy and inference time, we introduce an evaluation metric which we called \textit{Weighted Accuracy Inference Time Trade-off} (WAITT): 

\begin{equation}
\text{WAITT} = \frac{A}{IT^{\gamma} + \alpha \cdot (1 - A)}
\end{equation}
\noindent
where $A$ is accuracy, $IT$ is inference time, and $\alpha$ and $\gamma$ are two hyperparameters that can be tuned to penalize accuracy or inference time more. In this work, we set $\alpha$ to 1 and $\gamma$ to 2.

\begin{table}[t]
\centering
\resizebox{\textwidth}{!}{%
\begin{tabular}{lcccccc}
\toprule
\textbf{Model}          & \textbf{Input} & \textbf{Accuracy $\uparrow$} & \textbf{1 CS IIT(s) $\downarrow$} & \textbf{10 CS IIT(s) $\downarrow$} & \textbf{AVG IIT(s) $\downarrow$} & \textbf{WAITT $\uparrow$} \\ 
\midrule
EfficientNetV2         & Full RGB         & 0.726 & \textbf{3.59}  & \textbf{3.60}  & \textbf{0.001} & 2.649 \\
EfficientNetV2         & RGB Patches & 0.792 & 5.74  & 5.83  & 0.01   & 3.805 \\
EfficientNetV2         & Full Depth       & 0.778 & 9.04  & 10.02 & 0.108  & 3.329 \\
EfficientNetV2         & Depth Patches & \textbf{0.837} & 13.64  & 15.23 & 0.176  & \textbf{4.314} \\
\bottomrule
\end{tabular}%
}
\caption{Comparison between all EfficientNetV2 configurations.
All CNNs have been pretrained on ImageNet1K \cite{russakovsky2015imagenetlargescalevisual}.
Each row presents Accuracy, 1 CS IIT(s), 10 CS IIT(s), AVG IIT(s) (Average Inference Time per sample), and WAITT (Weighted Average Inference Time Trade-off).}
\label{tab:itime}
\end{table}

From Table \ref{tab:itime}, processing full RGB images achieves the fastest average inference time (0.001s), at the cost of lower accuracy (0.726), which negatively impacts the WAITT score, yielding the lowest WAITT value of 2.649. In contrast, extracting the athlete patch and passing it to the CNN instead of the entire image improves accuracy to 0.792, with a slightly higher latency (0.01s), resulting in the second-best WAITT score of 3.805. Using depth images leads to a better accuracy of 0.778, but the inference time increases significantly to 0.108s. The best WAITT score of 4.314 is achieved by using depth patches, resulting in the highest accuracy of 0.837 at the cost of a higher inference time (0.176s), making this the best option in contexts where low latency is not required.

\subsection{Comparison with a Skeleton-based Approach}
Finally, we compare the best-performing approaches with a skeleton-based method from \cite{finocchiaro2024calisthenics}, which uses OpenPose with the Body25B model to detect up to 25 joints. Each joint is represented by normalized (x, y) coordinates and a confidence score.  Detection is constrained to one person per frame. The classifier is an MLP with 75 input features (joint coordinates and confidence scores for 25 joints), and three hidden blocks, each with a linear layer of 512 neurons followed by a Leaky ReLU activation. The output layer uses a softmax function over 10 classes, representing the skills plus a background class. The model is trained using Adam (learning rate 0.001) with cross-entropy loss.

\begin{table}[t]
\centering
\resizebox{\columnwidth}{!}{%
\begin{tabular}{lcccccccc}
\toprule
\textbf{Approach} & \textbf{Input} & \textbf{Prec.}$\uparrow$ & \textbf{Rec.}$\uparrow$ & \textbf{F1}$\uparrow$ & \textbf{Acc.}$\uparrow$ & \textbf{AVG IIT (s)$\downarrow$} & \textbf{WAITT}$\uparrow$ & \textbf{Size (MB)$\downarrow$} \\
\midrule
\small{ENV2} & \small{RGB Patches} & 0.801 & 0.787 & 0.790 & 0.792 & \textbf{0.01} & 3.805 & \textbf{1,663} \\
\small{ENV2} & \small{Depth Patches} & \textbf{0.849} & \textbf{0.837} & \textbf{0.838} & \textbf{0.837} & 0.176 & \textbf{4.314} & 2,659 \\
OP-MLP & Full RGB & 0.812 & 0.815 & 0.813 & 0.815 & $\approx$ 0.383 & 2.457 & $\approx$ 2,617 \\
\bottomrule
\end{tabular}%
}
\caption{Comparison between the best-performing configurations of EfficientNetV2 (pretrained on ImageNet1K \cite{russakovsky2015imagenetlargescalevisual}) and the OP-MLP approach \cite{finocchiaro2024calisthenics}. For each row, we evaluated the approach using Precision, Recall, F1 Score, Accuracy, AVG IIT(s), WAITT, and Size(MB). ENV2: EfficientNetV2, OP: OpenPose, MLP: Multilayer Perceptron.}
\label{rgbvssb}
\end{table}

As shown in Table \ref{rgbvssb}, the EfficientNetV2 configuration with depth patches outperforms both the RGB patches and skeleton-based models across all metrics. It has a moderate inference time (0.176s per image), the highest WAITT score (4.314), and the second-highest memory usage (2,659MB).
The skeleton-based OP-MLP approach is the slowest (0.383s), with the lowest WAITT (2.457) and similar memory use ($\approx$2,617MB), ranking third in prediction performance.
The RGB patch approach is the fastest (0.01s per frame), although it has slightly worse prediction performance. It achieves the second-best WAITT score of 3.805 and has the lowest memory footprint (1,663MB), making it ideal for resource-constrained applications.

\subsection{Feature Maps Inspections}
In this section, we analyze EfficientNetV2's feature maps to understand the network's focus areas when processing RGB and depth images. Each feature map corresponds to a specific filter, with values indicating how strongly the filter detects patterns in the input.

\begin{figure}[t]
\includegraphics[width=\textwidth]{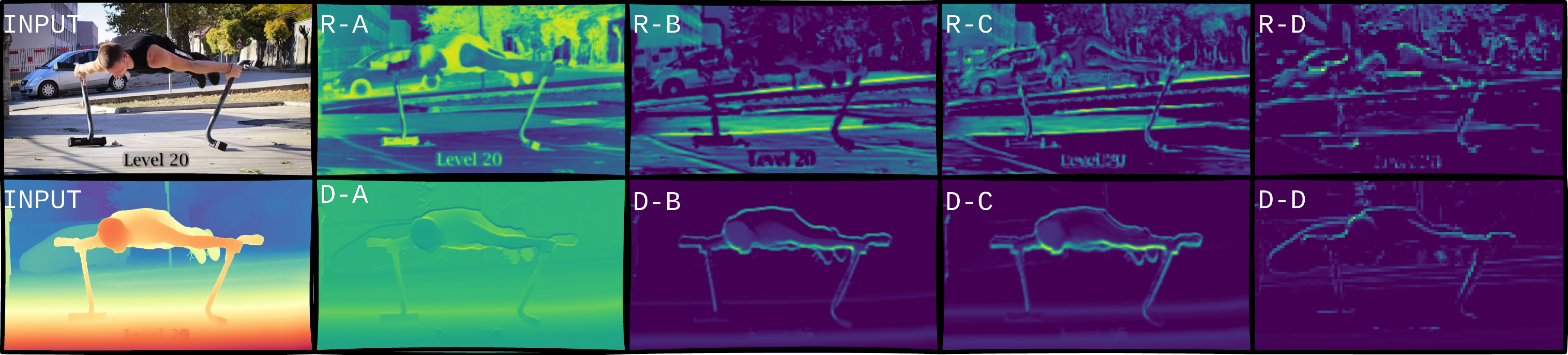}
\caption{Left: input image (RGB \& Depth), L: layer, B: block. The depth image propagates the athlete's information more smoothly, isolating him from the background features.}
\label{fig:fm}
\end{figure}

Figure \ref{fig:fm} shows feature maps from different network blocks for both RGB and depth inputs. The initial convolutional layer identifies basic features such as edges and textures, while subsequent layers develop more abstract representations of object parts and patterns. We specifically examine the first layers of blocks 2, 4, and 6 to demonstrate the progressive refinement of the network features. The comparison shows that RGB inputs introduce noise from background elements in deeper layers, while depth images maintain clearer, more focused representations of the foreground instance, with this advantage being particularly evident in the deeper blocks.

\section{Conclusion}
This work aimed to enhance image classification of calisthenics skills, providing a faster and more reliable alternative to skeleton-based pose classification \cite{finocchiaro2024calisthenics}. We explored four RGB-based approaches to improve performance.  
Classifying RGB images yielded the fastest inference time but the lowest accuracy. Using depth representations improved accuracy at the cost of slower inference. Introducing YOLOv10-based athlete detection and patch classification further enhanced performance, with depth patches achieving the highest accuracy. To better evaluate the trade-off between accuracy and inference time, we designed WAITT, a metric that balances both aspects. Our best-performing model, based on depth patches, achieved the highest WAITT score, outperforming the skeleton-based approach in efficiency.
While our approaches outperform skeleton-based methods in robustness, the task remains open. Advances in depth estimation and CNN models could further improve accuracy and speed. Our modular approaches allow flexible replacement of components, providing a scalable template for classification in different domains.

Code and weights can be found at \\ \url{https://github.com/antof27/rgb-based-pose-classification}.


\section*{Acknowledgements}
This research has been supported by Research Program PIAno di inCEntivi per la Ricerca di Ateneo 2020/2022 (C.d.A. del 29.04.2020) — Linea di Intervento 3 ``Starting Grant" - University of Catania.

%

%
%
%
%





\bibliographystyle{splncs04}
\bibliography{main}

\end{document}